%% file: main.tex
\documentclass[runningheads]{llncs}

\usepackage{eccvabbrv}

\usepackage{graphicx}
\usepackage{booktabs}
\usepackage{multirow}

\usepackage{algorithm}
\usepackage{algpseudocode}

\usepackage{cuted}
\usepackage{capt-of}

\usepackage[accsupp]{axessibility}  

\usepackage{hyperref}

\usepackage{orcidlink}

\begin{document}

\title{RoboClaw: An Agentic Framework for Scalable Long-Horizon Robotic Tasks} 

\titlerunning{RoboClaw}

\author{Ruiying Li\inst{1,2}\thanks{Equal contribution.}  \and
Yunlang Zhou\inst{1,3}$^*$ \and
Yuyao Zhu\inst{1,3} \and
Kylin Chen\inst{1} \and
Jingyuan Wang\inst{1} \and
Sukai Wang\inst{1} \and
Kongtao Hu\inst{1} \and
Minhui Yu\inst{1} \and
Bowen Jiang\inst{1} \and
Zhan Su\inst{1,3} \and
Jiayao Ma\inst{1} \and
Xin He \inst{1} \and
Yongjian Shen\inst{1} \and
Yang Yang\inst{1} \and
Guanghui Ren\inst{1} \and
Maoqing Yao\inst{1} \and
Wenhao Wang\inst{1}$^\dagger$ \and
Yao Mu\inst{3,4}$^\dagger$
}

\authorrunning{R. Li, Y. Zhou et al.}

\institute{AgiBot, China \and
National University of Singapore \and
Shanghai Jiao Tong University, Shanghai 200240, China \and
MoE Key Lab of Artificial Intelligence, AI Institute, SJTU
}

\maketitle
\begingroup
\renewcommand{\thefootnote}{\ensuremath{\dagger}}
\footnotetext{Corresponding authors. wangwenhao@agibot.com, muyao@sjtu.edu.cn}
\endgroup
\setcounter{footnote}{0}

\input{section/abstract}

\input{section/introduction}

\input{section/related_works}

\input{section/method}

\input{section/experiments}

\input{section/conclusion}

\bibliographystyle{splncs04}
\bibliography{main}
\end{document}

%% file: section/abstract.tex
\begin{figure}[t]
    \centering
    \includegraphics[width=0.95\textwidth]{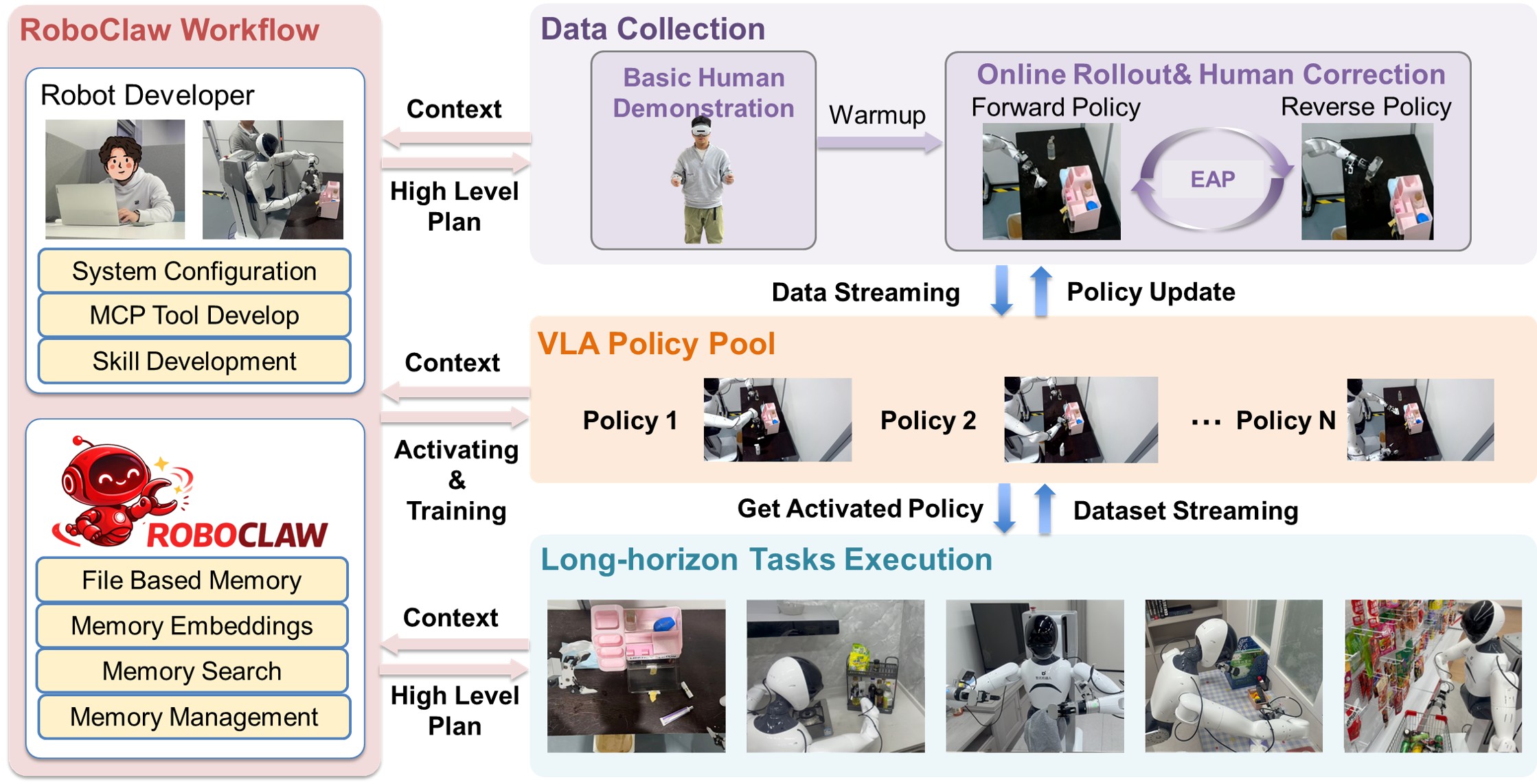}
    \caption{RoboClaw workflow across the robot policy lifecycle. A robot developer specifies system configuration, MCP tools, and skills, while RoboClaw provides file-based memory, memory embeddings, search, and management. Data is collected through basic human demonstrations followed by online rollout with EAP self resetting, producing a VLA policy pool that is continuously updated via streaming data. Activated policies are then used to execute complex long-horizon tasks under high-level plans and contextual guidance.}
    \label{fig:teaser}
\end{figure}

\begin{abstract}

Vision-Language-Action (VLA) systems have shown strong potential for language-driven robotic manipulation. However, scaling them to long-horizon tasks remains challenging. Existing pipelines typically separate data collection, policy learning, and deployment, resulting in heavy reliance on manual environment resets and brittle multi-policy execution.  We present \textbf{RoboClaw}, an agentic robotics framework that unifies data collection, policy learning, and task execution under a single VLM-driven controller. At the policy level, RoboClaw introduces \textbf{Entangled Action Pairs (EAP)}, which couple forward manipulation behaviors with inverse recovery actions to form self-resetting loops for autonomous data collection. This mechanism enables continuous on-policy data acquisition and iterative policy refinement with minimal human intervention. During deployment, the same agent performs high-level reasoning and dynamically orchestrates learned policy primitives to accomplish long-horizon tasks. By maintaining consistent contextual semantics across collection and execution, RoboClaw reduces mismatch between the two phases and improves multi-policy robustness. Experiments in real-world manipulation tasks demonstrate improved stability and scalability compared to conventional open-loop pipelines, while significantly reducing human effort throughout the robot lifecycle, achieving a 25\% improvement in success rate over baseline methods on long-horizon tasks and reducing human time investment by 53.7\%. Code is available at: \href{https://github.com/RoboClaw-Robotics/RoboClaw}{https://github.com/RoboClaw-Robotics/RoboClaw}.
\end{abstract}

%% file: section/introduction.tex
\section{Introduction}
\label{sec:intro}


Recent advances in Vision-Language-Action (VLA) systems have demonstrated significant potential for language-driven robotic manipulation, enabling multimodal models to map language instructions and visual observations directly to robot actions \cite{blackP0VisionLanguageActionFlow,brohanRT2VisionLanguageActionModels,brohanRT1RoboticsTransformer2023,driessPaLMEEmbodiedMultimodal2023,kimOpenVLAOpenSourceVisionLanguageAction}. However, scaling this paradigm to complex, real-world manipulation tasks remains a critical challenge. Real-world robotic tasks are inherently long-horizon and compositional, requiring the sequential execution of multiple interdependent subtasks.
To this end, VLA systems commonly rely on large-scale robot data to learn diverse task policies. However, constructing such datasets in real robotic environments often requires substantial human involvement. Operators must collect demonstrations, repeatedly reset environments, monitor failures, filter trajectories, evaluate model performance, and supervise robot behavior during downstream long-horizon task execution. As task complexity grows, this human-centered data collection and deployment process becomes increasingly costly and difficult to scale. Moreover, these stages are often handled by different individuals, introducing information gaps across the system pipeline. As a result, the interpretation of task states, subtask boundaries, or success criteria may differ across stages, making it difficult to maintain consistent task semantics throughout the system.

Furthermore, when data collection, model learning, and task execution are driven by independent processes, the state distribution covered by training data often fails to reflect the conditions encountered during deployment, leading to a mismatch between training and execution. Such inconsistencies in both semantics and distribution make long-horizon tasks particularly brittle, where small errors may propagate and cascade through the execution process. Therefore, a key challenge is how to establish a unified semantic representation and decision mechanism across data collection, policy learning, and execution for scalable language-driven robotic systems.

To address this problem, we introduce \textbf{RoboClaw}, a unified agent architecture for long-horizon robotic manipulation, as illustrated in Figure~\ref{fig:teaser}. RoboClaw follows a simple interaction paradigm: a user sends a task instruction, and the robot autonomously reasons and executes the task. In this framework, a Vision-Language-Model (VLM) acts as a meta-controller that performs high-level decision making through in-context learning (ICL)\cite{dong2024surveyincontextlearning}, reasoning over both environmental observations and structured memory. Unlike traditional systems that rely on manual supervision or predefined planners, RoboClaw unifies data collection, policy learning, and task execution within a single agent loop, enabling consistent task semantics and decision logic throughout the system lifecycle and shifting robotic operation from human-gated operation toward agentic operation.

At the data acquisition stage, RoboClaw introduces \textbf{Entangled Action Pairs (EAP)}, a mechanism that significantly reduces the need for manual environment resets. For each manipulation policy, we pair a forward execution behavior with a complementary inverse recovery behavior, forming a self-resetting loop that allows the robot to repeatedly return to a reusable precondition region. Under agent control, these paired actions alternate execution, enabling continuous online data collection without frequent human intervention. Compared with traditional pipelines that rely on manual resets or demonstrations, this mechanism substantially reduces human effort while maintaining alignment between collected data and execution conditions.

During task execution, RoboClaw also relies on the agent to orchestrate skill invocation. Rather than following static skill sequences or requiring constant human monitoring, the agent dynamically selects and schedules modular skills based on the current context. By continuously monitoring subtask states and validating execution conditions, the agent performs runtime supervision and triggers recovery behaviors when necessary, leading to a 25\% higher success rate on long-horizon tasks compared to baseline approaches.

Finally, RoboClaw establishes a closed-loop lifecycle learning mechanism. Execution trajectories generated during downstream long-horizon task execution can be reintegrated into the training pipeline under the same contextual semantics and decision policy, enabling continual improvement of existing policies and expansion of the policy pool. By unifying data acquisition, model learning, and task execution within a single agent framework, the system can accumulate experience and improve performance over time. When abnormal situations or safety constraints are detected, the system can also request human intervention, ensuring operational safety while reducing human burden by 53.7\%.

Our contributions are summarized as follows:

\textbf{A lifecycle agentic framework for robotics.}  
We introduce RoboClaw, an agentic framework that unifies data collection, policy learning, and long-horizon task execution, enabling consistent contextual semantics and significantly reducing human burden.

\textbf{Learning-driven autonomous data collection.}  
We propose Entangled Action Pairs (EAP), a data engine that couples forward manipulation polices with inverse behaviors to form self-resetting loops, enabling continuous online data collection and maintaining alignment between collected data and execution conditions.

\textbf{Skill orchestration and status monitoring for long-horizon tasks.}  
We design a context-driven decision architecture where a VLM performs high-level reasoning through in-context learning over structured memory, enabling skill orchestration and state monitoring for long-horizon robotic manipulation.

%% file: section/related_works.tex
\section{Related Works}

\subsection{Closed-loop Data Collection}

Recent methods explore closed-loop and semi-automated pipelines to scale robot learning, moving beyond standard teleoperation systems like AnyTeleop~\cite{qinAnyTeleopGeneralVisionBased2023}, GELLO~\cite{wuGELLOGeneralLowCost2024}, and Mobile ALOHA~\cite{fuMobileALOHALearning2024}. To reduce human burden in the real world, systems like RoboCopilot~\cite{wu2025robocopilothumanintheloopinteractiveimitation} utilize human-in-the-loop residual corrections. Genie Centurion~\cite{wang2025genie} introduces a ``rewind-and-refine'' mechanism guided by a Task Sentinel that autonomously detects failures to request human intervention, and VLAC~\cite{zhai2025vision} has a similar mechanism. Furthermore, FieldGen~\cite{wang2025fieldgen} semi-automates real-world collection by decoupling manipulation phases, using human demonstrations only for fine manipulation while automatically synthesizing diverse pre-manipulation trajectories via attraction fields.

To fully automate data collection, systems like MimicGen~\cite{wangMimicPlayLongHorizonImitation2023}, GenH2R-Sim~\cite{wangGenH2RLearningGeneralizable2024}, and RoboCasa~\cite{nasirianyRoboCasaLargeScaleSimulation2024} synthesize large-scale demonstrations in simulation. Recent advances also leverage Large Language Models (LLMs) for task planning and automated execution. For instance, RoboTwin 2.0~\cite{chenRoboTwin20Scalable2025} employs MLLMs with simulation-in-the-loop feedback to iteratively validate and refine task execution code. HumanoidGen~\cite{jingHumanoidGenDataGeneration2025} leverages LLMs to generate spatial constraints for humanoid manipulation and employs an STCR-based tree search mechanism to improve long-horizon task planning. Additionally, CyberDemo~\cite{wangCyberDemoAugmentingSimulated2024} introduces a learning-driven closed-loop via Auto Curriculum Learning, dynamically adjusting data augmentation complexity based on the policy's current success rate.

While these works successfully integrate closed-loop feedback for data synthesis or human-assisted correction, they often lack autonomous adaptability during real-world deployment. To address this gap, our work proposes a fully learning-driven automated data collection framework. Crucially, unlike systems requiring manual intervention or predefined fields, we introduce autonomous process monitoring and skill scheduling during inference. This enables real-time error recovery and robust execution in dynamic environments without human assistance.

\subsection{Foundation Models for Embodied Tasks}
In recent years, Vision-Language-Action (VLA) models such as PaLM-E~\cite{driessPaLMEEmbodiedMultimodal}, RT-2~\cite{brohanRT2VisionLanguageActionModels}, OpenVLA~\cite{kimOpenVLAOpenSourceVisionLanguageAction}, and $\pi_0$~\cite{blackP0VisionLanguageActionFlow} have advanced language-conditioned robotic control by unifying perception, language, and action, yet remain susceptible to error accumulation in long-horizon tasks. Large language models have also been adopted for planning, including Language Models as Zero-Shot Planners~\cite{huangLanguageModelsZeroShot}, Code as Policies~\cite{liangCodePoliciesLanguage2023}, and VoxPoser~\cite{huangVoxPoserComposable3D}, improving task decomposition but offering limited execution-time supervision. Hierarchical approaches such as SayCan~\cite{ahnCanNotSay}, HAMSTER~\cite{liHAMSTERHierarchicalAction2025}, HiRobot~\cite{shiHiRobotOpenEnded2025}, and Agentic Robot~\cite{yangAgenticRobotBrainInspired2025} introduce structured subtask abstraction and plan--verify mechanisms, while $\pi_{0.5}$~\cite{blackP05VisionLanguageActionModel} strengthens multi-stage reasoning within a unified VLA framework. Inner Monologue~\cite{huangInnerMonologueEmbodied} and LITEN~\cite{shahLearningAffordancesInferenceTime2025} enhance robustness through replanning, yet sustained process-level supervision during execution remains largely unexplored.

In contrast, we propose a context-aware supervisory agent operating at inference time to continuously monitor subtask execution and dynamically select retry, recovery, or human intervention strategies. Decoupled from specific task structures or skill libraries, our design enables scalable real-world long-horizon embodied agents.

%% file: section/method.tex
\begin{figure}[t]
    \centering
    \includegraphics[width=0.95\linewidth]{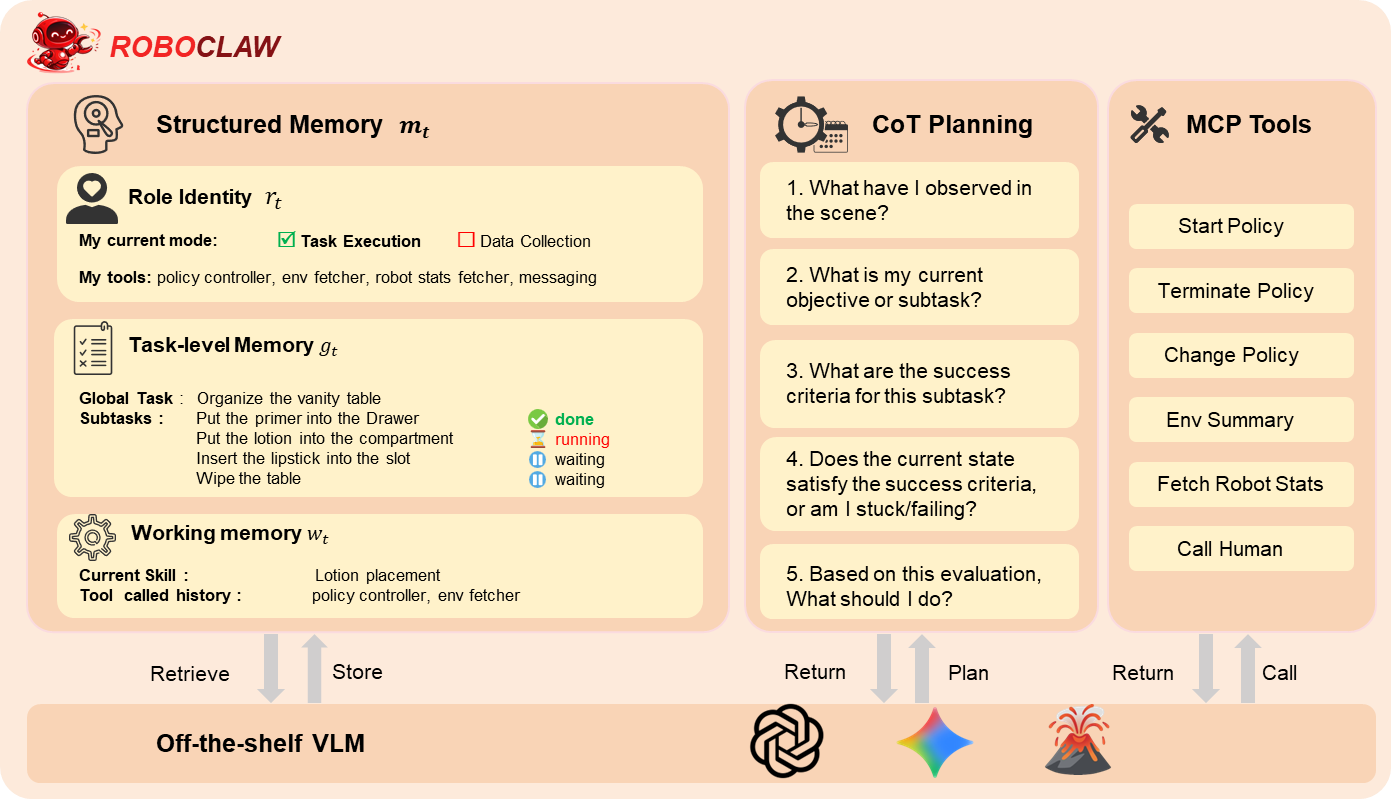}
    \caption{
        RoboClaw system architecture.
A Vision-Language-Model (VLM) acts as a meta-controller operating under an in-context learning paradigm.
Multimodal observations are integrated with structured memory
(role identity, task-level memory, and working memory) to form the decision context.
Through chain-of-thought (CoT) reasoning, the agent generates high-level decisions and invokes tools through a unified MCP execution interface.
The same agent core governs both data collection and policy deployment,
ensuring consistent control semantics across the full system lifecycle.
    }
    \label{fig:agent_mechanism}
\end{figure}

\begin{figure}[t]
    \centering
    \includegraphics[width=0.95\linewidth]{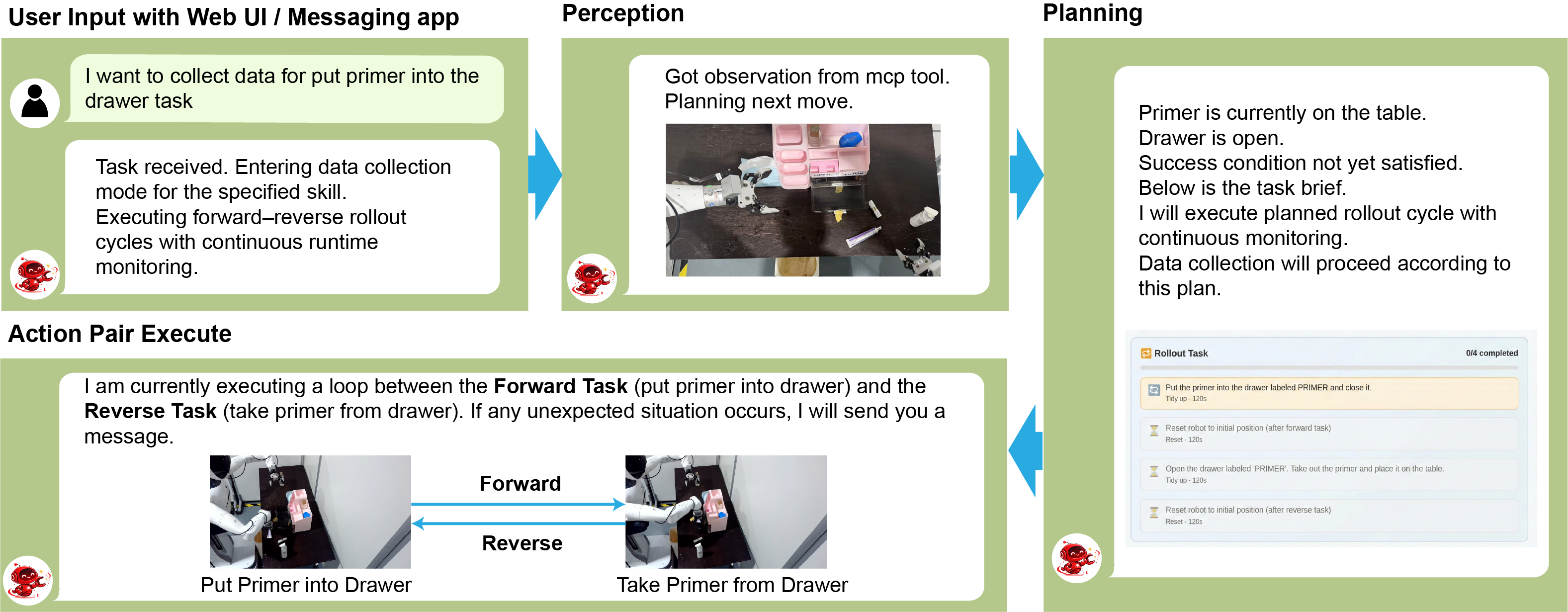}
    \caption{
    RoboClaw Autonomous Data Collection Workflow. This diagram illustrates the process of the agent interacting with a user to initiate a data collection task for the robot ("place the primer into the drawer"). The agent autonomously processes visual observations using MCP tools, evaluates the initial state of the environment, and formulates a task plan. Subsequently, it continuously executes a forward-reverse operational loop (i.e., placing the item into the drawer and then taking it out) while monitoring for anomalies in real-time during execution, thereby continuously acquiring the robotic manipulation dataset.
    }
    \label{fig:eap_collection}
\end{figure}

\section{Method}
\label{sec:method}

We propose \textbf{RoboClaw}, an agentic framework for long-horizon robotic manipulation that unifies autonomous data collection and task execution. 
RoboClaw employs an off-the-shelf Vision–Language-Model (VLM) as a high-level controller that reasons over visual observations and system context to decide which skill to invoke.

The system operates in a closed-loop agent interaction cycle. 
Given observations and structured memory, the VLM performs chain-of-thought (CoT) reasoning to interpret the current task state, evaluate progress, and determine the next action. 

RoboClaw integrates structured memory with a modular skill library in an OpenClaw style, enabling the agent to compose reusable capabilities for complex workflows such as data collection and task execution. We structure the system into three hierarchical levels of abstraction: \textit{Skills}, \textit{Tools}, and \textit{Policies}, where higher levels invoke the lower levels to accomplish tasks. We introduce these three components in reverse order. \textit{Policies} refer to a robotic foundation model that produces low-level motor actions, implemented as Vision–Language–Action (VLA) models in our system. \textit{Tools} are callable system interfaces (e.g., Start Policy, Terminate Policy, Env Summary) that allow the agent to execute policies or query the environment through the Model Context Protocol (MCP). \textit{Skills} denote reusable procedures that orchestrate tools, e.g., a ``long-horizon-execution'' skill may call Env Summary and then call Start Policy to execute manipulation policy.

\subsection{Autonomous Robotic Task Execution and Data Collection via RoboClaw Agentic Framework}
\label{sec:framework}
Building on the hierarchical design described above, we now introduce the overall execution framework that enables RoboClaw to perform autonomous task execution and data collection.

As illustrated in Fig.~\ref{fig:agent_mechanism}, RoboClaw organizes the agent's perception, reasoning, and action into a closed-loop decision process that iteratively updates memory and interacts with the environment.

At each timestep $t$, the agent maintains a structured memory state $m_t$ that provides contextual information for reasoning and planning. The memory consists of three components. The \emph{role identity} $r_t$ specifies the current operational mode of the agent and the set of available tools. The \emph{task-level memory} $g_t$ records the global task together with its decomposed subtasks and their execution status, enabling the agent to track long-horizon task progress. The \emph{working memory} $w_t$ stores short-term execution context such as the currently active skill and the history of tool invocations. During execution, the agent continuously retrieves and updates this structured memory.

Given the current observation and memory state, the VLM performs structured reasoning through a chain-of-thought (CoT) planning process. The reasoning procedure first interprets the current scene and identifies the relevant elements in the environment. It then determines the current objective or subtask and evaluates the criteria for successful completion. Based on this evaluation, the agent assesses whether the current state satisfies the task requirements or whether corrective actions are needed, and finally decides the next action to execute.

To bridge high-level reasoning with robot control, the framework provides a set of external tools through a Model Context Protocol (MCP) interface. These tools allow the agent to start, terminate, or switch control policies, retrieve environment summaries, query robot states, and request human intervention when necessary. By invoking these tools, the agent translates high-level plans generated by the VLM into executable actions.

Overall, the agent operates in an iterative loop: it retrieves relevant information from structured memory and environment observations, performs CoT-based reasoning to determine the next action, and executes the corresponding tool call. The resulting outcomes are written back into memory, forming a continuous perception–reasoning–action cycle until the task is completed.

\subsection{Self-Resetting Data Collection via Entangled Action Pairs}
\label{sec:data_collection}

\begin{figure}[t]
    \centering
        \includegraphics[width=0.95\linewidth]{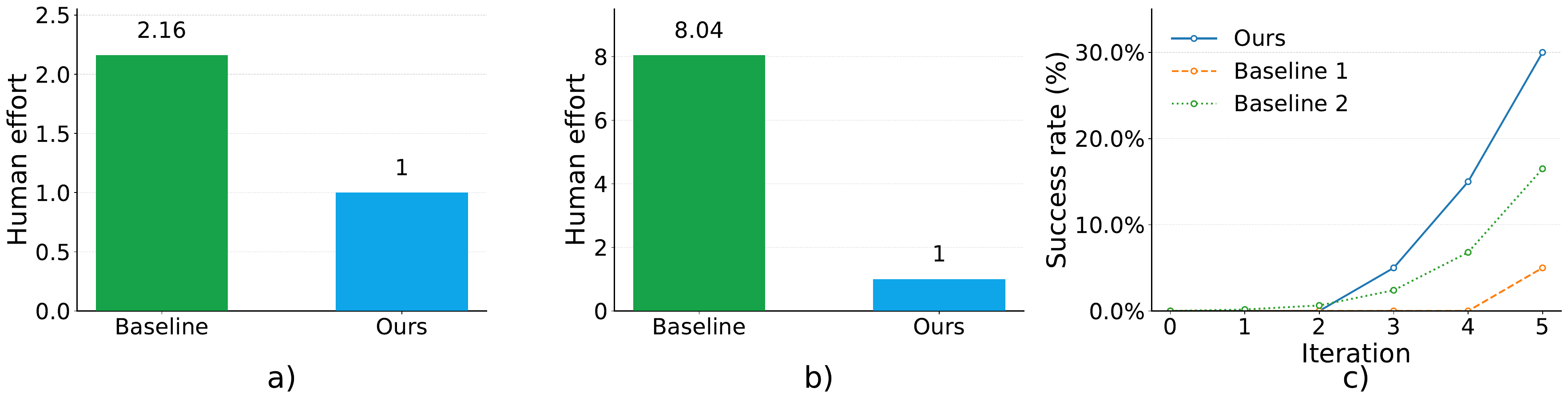}
    \caption{
Human effort comparison for data collection.
(a) Relative human time required to collect the same amount of data.
(b) Relative human intervention during rollout execution.
All values are normalized with respect to our method (Ours = 1).
(c) Success rate across iterations on the vanity table organization task. RoboClaw (Ours) significantly outperforms both end-to-end VLA baselines and the expected success rate computed as the product of four independent subtask success rates. The improvement comes from RoboClaw’s ability to monitor task progress and automatically invoke recovery policies when failures occur.
Results are averaged over 20 trials.
}
    \label{fig:human_effort}
\end{figure}

During data collection, RoboClaw operates as a data collector and interacts
with the environment within a closed loop consisting of structured memory,
a CoT planning module, and tool interfaces (see Fig.~\ref{fig:agent_mechanism}).
At time step $t$, the agent receives visual observation $o_t$ and maintains
a structured memory state:
\begin{equation}
m_t = (r_t, g_t, w_t)
\end{equation}
where $r_t$ denotes the role identity of RoboClaw, $g_t$ is the task-level
memory that records the global task and subtask progress, and $w_t$ is the
working memory that stores the currently activated skill and the history
of tool invocations. Through the observation and memory, the agent can
reason about the current scene and the task execution status.

The agent selects the next subtask $z_t$ from the candidate subtask set
$\mathcal{Z}$:
\begin{equation}
z_t = \operatorname{RoboClaw}(m_t, o_t), \quad z_t \in \mathcal{Z}.
\end{equation}
The agent evaluates whether the subtask has been successfully completed and updates the task memory $g_t$ accordingly.

The low-level manipulation policies in RoboClaw are implemented using the
Vision-Language-Action (VLA) model $\pi_{0.5}$~\cite{blackP05VisionLanguageActionModel}.
VLA policies jointly process visual observations, language instructions,
and robot proprioceptive states to generate executable robot actions.

In our system, the language instruction is not directly provided by a
human operator. Instead, it is dynamically generated by the RoboClaw agent
during MCP tool invocation. When RoboClaw decides to execute a skill,
it produces a structured instruction describing the current subtask,
which is used to condition the policy.

Formally, the policy predicts a short-horizon action sequence
\begin{equation}
A_t = \pi_{0.5}(o_t, l_t, q_t),
\end{equation}
where $o_t$ denotes the visual observation, $l_t$ denotes the instruction
generated by the RoboClaw agent, and $q_t$ denotes the robot joint state.
The predicted action chunk (with length $H$) is defined as
\begin{equation}
A_t = [a_t, \ldots, a_{t+H-1}],
\end{equation}
The policy is trained to model the distribution $p(A_t \mid o_t,l_t,q_t)$ using a conditional flow matching objective. It learns a velocity field $v_\theta$ that transports a standard Gaussian noise distribution to the true action distribution.
\begin{equation}
\mathcal{L}^{\tau}(\theta)
=
\mathbb{E}_{p(A_t \mid o_t,l_t,q_t),\,q(A_t^\tau \mid A_t)}
\left[
\left\|
v_\theta(A_t^\tau,o_t,l_t,q_t)
-
u(A_t^\tau \mid A_t)
\right\|^2
\right],
\end{equation}
where $\tau\in[0,1]$ is the flow matching time step, and $A_t^\tau = (1-\tau)\epsilon + \tau A_t$ is the linearly interpolated state between the sampled Gaussian noise $\epsilon$ and the ground-truth action chunk $A_t$.

For each policy $k$, we learn a forward execution policy
$\pi^{\rightarrow}_{\theta_k}$ and a reset policy
$\pi^{\leftarrow}_{\phi_k}$.
The forward interaction collects a trajectory
\begin{equation}
\tau_k^{\rightarrow}
=
\{(o_t, q_t, a_t)\}_{t=0}^{T},
\qquad
a_t = \pi^{\rightarrow}_{\theta_k}(o_t, l_t, q_t).
\end{equation}
Once the agent determines that the subtask has been successfully completed,
the reset policy is triggered to restore the environment state:
\begin{equation}
\tau_k^{\leftarrow}
=
\{(o_t, q_t, a'_t)\}_{t=T+1}^{T+T_{\text{reset}}},
\qquad
a'_t = \pi^{\leftarrow}_{\phi_k}(o_t, l_t, q_t).
\end{equation}
These two trajectories together form an entangled pair
\begin{equation}
\tau_k = (\tau_k^{\rightarrow}, \tau_k^{\leftarrow}),
\end{equation}
enabling the environment to automatically return to its initial state
without human intervention. All collected trajectories are stored in the
dataset $\mathcal{D}$ for subsequent policy learning.

\subsection{Deployment-time Process Supervision and Skill Scheduling}
\label{sec:deployment}

During deployment, RoboClaw operates as a task executor and composes
previously learned policies to accomplish long-horizon tasks.
The execution follows the same closed-loop decision structure introduced
in Sec.~\ref{sec:data_collection}, where the agent reasons over the
current observation $o_t$ and structured memory $m_t$ to select the next
subtask $z_t$.

Given the selected subtask, RoboClaw invokes the corresponding forward
policy from the forward policy set
$\{\pi^{\rightarrow}_{\theta_k}\}_{k=1}^{K}$ through the MCP tool interface.
During execution, the agent periodically queries environment summaries
and robot status (e.g., via Fetch Robot Stats and
Env Summary) to monitor task progress.
These feedback signals are written into working memory $w_t$ and used to
evaluate whether the current subtask has been completed.

If the success condition of the subtask is satisfied, the agent updates
the task-level memory $g_t$ and proceeds to the next subtask in the task plan.
Otherwise, the agent may retry the same policy or switch to another
forward policy via the Change Policy tool.

If the system detects repeated failures or unexpected environment states,
RoboClaw attempts recovery by re-planning and selecting alternative skills
from the forward skill set.
When autonomous recovery is unsuccessful or safety conditions are triggered,
the agent escalates to human intervention through the Call Human
tool via the MCP interface.
This design allows the system to operate autonomously in most cases while
preserving human oversight for safety-critical situations.

Importantly, the trajectories generated during deployment are recorded
and incorporated into the dataset $\mathcal{D}$.
These trajectories capture additional state distributions encountered
during real task execution and can be used to further refine the skill
policies $\{\pi^{\rightarrow}_{\theta_k}\}$.
In this way, deployment not only executes tasks but also serves as an
additional source of experience for improving the skill library.

By sharing the same decision loop and skill interface across both
data collection and deployment, RoboClaw forms a unified lifecycle
learning framework in which execution continuously improves the
underlying skills.

\input{tables/train_config_table}

%% file: tables/train_config_table.tex


\begin{table}[t]
\centering
\caption{Training hyperparameters for $\pi_{0.5}$ fine-tuning.}
\label{tab:train_config}
\setlength{\tabcolsep}{4pt}
\renewcommand{\arraystretch}{1.05}
\begin{tabular}{llll}
\toprule
\multicolumn{2}{c}{\textbf{General Settings}} & 
\multicolumn{2}{c}{\textbf{LoRA Settings}} \\
\cmidrule(r){1-2} \cmidrule(l){3-4}
Precision & bfloat16 &
Rank ($r$) & 16 \\
Batch size & 16 &
Alpha ($\alpha$) & 16 \\
Training steps & 10k &
Dropout & 0.1 \\
Warmup steps & 100 &
Target modules & all-linear \\
Learning rate & $2.5\times10^{-5}$ &
Inference steps & 3 \\
Gradient checkpointing & \checkmark & & \\
\bottomrule
\end{tabular}
\end{table}

%% file: section/experiments.tex
\input{tables/subtask_reset}

\input{tables/subtask_effiency}

\section{Experiments}
\label{sec:experiments}

RoboClaw is designed to address four key challenges faced by robotic manipulation systems in real-world environments: \textbf{(1)} improving data collection efficiency, \textbf{(2)} increasing the success rate of subtask policies, \textbf{(3)} improving performance on complex long-horizon tasks, and \textbf{(4)} learning from failures. 

To evaluate the capabilities of RoboClaw, we design a set of real world manipulation tasks as our experimental scenarios. All experiments are conducted on the Agibot G01 platform, a dual-arm mobile manipulation robot mounted on a mobile base. The platform provides 20 degrees of freedom excluding the end-effectors and each arm is equipped with an AGIBOT OmniPicker gripper, an adaptive gripper with a single active degree of freedom.

Based on this experimental platform, our evaluation focuses on the following four key questions:
\begin{enumerate}
    \item Can RoboClaw significantly improve data collection efficiency?
    \item Can RoboClaw improve the success rate of subtask policies?
    \item Can RoboClaw improve the robot's performance on complex long-horizon tasks?
    \item Can RoboClaw learn from failures?
\end{enumerate}

\subsection{Can RoboClaw Improve Data Collection Efficiency?}
\label{sec:exp_efficiency}

To answer Question~(1), we evaluate RoboClaw in four real-world scenarios: a bedroom vanity table, a kitchen shelf, a study desk, and a convenience-store shelf. These environments represent common organization and retrieval tasks in both household and retail settings, providing a diverse set of manipulation challenges.

In each environment, the robot is assigned a corresponding organization or retrieval task. For example, in the bedroom scenario, the robot organizes items on a vanity table; in the kitchen, it arranges objects on a storage shelf; in the study, it tidies items on a desk; and in the convenience-store scenario, it selects specific products according to given instructions. These tasks typically require manipulating multiple objects and executing a sequence of actions, and therefore fall into the category of \textit{multi-stage manipulation tasks}. In addition, correct interpretation of task instructions and semantic identification of target objects are also used as criteria for determining task success.

We compare our approach with a baseline that relies on purely manual data collection, where human operators perform demonstrations and manually reset the environment after each trial. Given the same amount of collected data, we measure the proportion of human effort required by each method.

As shown in Fig.~\ref{fig:human_effort}(a), the RoboClaw data collection pipeline
substantially reduces the amount of human time required to obtain the same number of trajectories.
When normalized by the human effort required by our method (Ours = 1),
the manual data collection baseline requires approximately $2.16\times$ more human time.

We further analyze the fraction of human intervention during model rollouts in RoboClaw pipeline.
As illustrated in Fig.~\ref{fig:human_effort}(b), the manual baseline requires frequent human involvement,
while RoboClaw performs most data collection autonomously.
The baseline requires approximately $8.04\times$ more human intervention compared to our method.

The results indicate that RoboClaw consistently improves data collection efficiency across all tested environments. Compared to traditional manual demonstrations, RoboClaw can autonomously monitor the environment state and repeatedly perform \textit{Entangled Action Pairs} to continuously generate new trajectories.

Overall, this capability for autonomous data collection in real-world environments substantially reduces reliance on human demonstrations and significantly lowers both the labor cost and time cost of data acquisition. As a result, RoboClaw provides a much more efficient approach to generating large-scale training data for robotic learning systems.

\subsection{Can RoboClaw Improve the Success Rate of Subtask Policies?}
\label{sec:exp_quality}

In the next set of experiments, we investigate whether RoboClaw can improve the success rate of four individual subtask policies. In our setup, the storage organizer contains compartments labeled with object categories, and the robot must interpret these labels and place objects into the corresponding compartments. The four single-skill tasks are intentionally different. One is mostly about long-range pick-and-place, one adds a constrained follow-up interaction, one is a tight insertion problem, and one depends on sustained surface contact. Put together, they give a reasonable spread of manipulation difficulty without making the setup hard to interpret.

  \textbf{Body Lotion placement.} The robot needs to pick up a bottle of body lotion from the vanity table and move it to the labeled placement
  area. This task is challenging due to the large amount of motion involved. The bottle travels across a fairly large part of the
  workspace, and the camera view changes a lot between approach, grasp, lift, and placement.

\textbf{Primer placement.} 
The robot needs to place a tube of primer into a target region inside the labeled drawer and then close the drawer. 
The additional closing step increases the difficulty of the task, as successful execution requires not only accurate object placement but also leaving the scene in a state that allows the drawer to be closed reliably. 
The drawer further complicates perception and placement due to occlusion and limited clearance. 

\textbf{Lipstick insertion.} 
The robot needs to insert a lipstick into the labeled narrow slot. 
This task involves tight positional and rotational tolerances, making accurate alignment critical for successful insertion. 
Even small deviations during execution can lead to failure at the insertion stage. 
Therefore, the policy must maintain precise alignment with the slot until contact to ensure successful insertion.

\textbf{Tissue wipe.} 
The robot needs to use a tissue to wipe a designated region on the table containing spilled toning water. 
Unlike the previous tasks, success in this task depends on the quality of a continuous motion rather than achieving a single final pose. 
The robot must maintain stable contact with the surface and execute a consistent wiping trajectory that sufficiently covers the target area. 
Loss of contact or unstable motion can lead to ineffective wiping and task failure.

Even with sufficient training data, robot policies may still fail due to the inherent difficulty of the tasks, environmental variations, or execution errors. Improving the reliability and robustness of individual policies is therefore critical for building dependable robotic systems.

In this experiment, we analyze the effect of iterative data collection by varying the number of iterations in the RoboClaw pipeline. Specifically, we train models using data collected from one to five iterations, where each iteration adds 50 additional data samples. We then compare their performance to evaluate how iterative rollout affects policy success rates.

During policy training, we allocate a fixed number of human demonstrations for each forward policy.
Additional training data are obtained through the closed-loop rollout process described in Section~\ref{sec:data_collection}.

We evaluate the performance of the inverse reset policies used during the data collection loop.
As shown in Table~\ref{tab:inverse_skill}, the inverse policies achieve relatively high success rates across all tasks,
with the number of successful trials ranging from 36/50 to 43/50.
This is expected because, in order to enable automatic data collection for the more challenging forward tasks,
the inverse tasks are intentionally designed to be simpler than the forward tasks themselves.

Table~\ref{tab:forward_skill} reports the success rates of forward policies across rollout iterations.
As additional rollout trajectories are incorporated into the training set, the performance of all policies improves steadily.
For example, the success rate of the \textit{Body Lotion} policy increases from 21/50 in the first iteration to 43/50 in the fifth iteration,
while the \textit{Primer} policy improves from 23/50 to 40/50.
More challenging manipulation tasks also benefit from iterative rollout:
the success rate of the \textit{Lipstick} insertion policy increases from 2/50 to 23/50,
and the \textit{Tissue Wipe} policy improves from 11/50 to 26/50.
These results indicate that closed-loop collected trajectories provide more informative training data
and improve the robustness of individual policies under the same human demonstration budget.

Notably, the asymmetry between forward and inverse policies is beneficial for the EAP data collection mechanism.
Reliable inverse policies help maintain stable self-resetting loops,
allowing the robot to continuously collect trajectories with minimal human intervention.

\subsection{Can RoboClaw Better Handle Long-Horizon Tasks?}
\label{sec:exp_longhorizon}

\begin{figure}[t]
    \centering
        \includegraphics[width=0.95\linewidth]{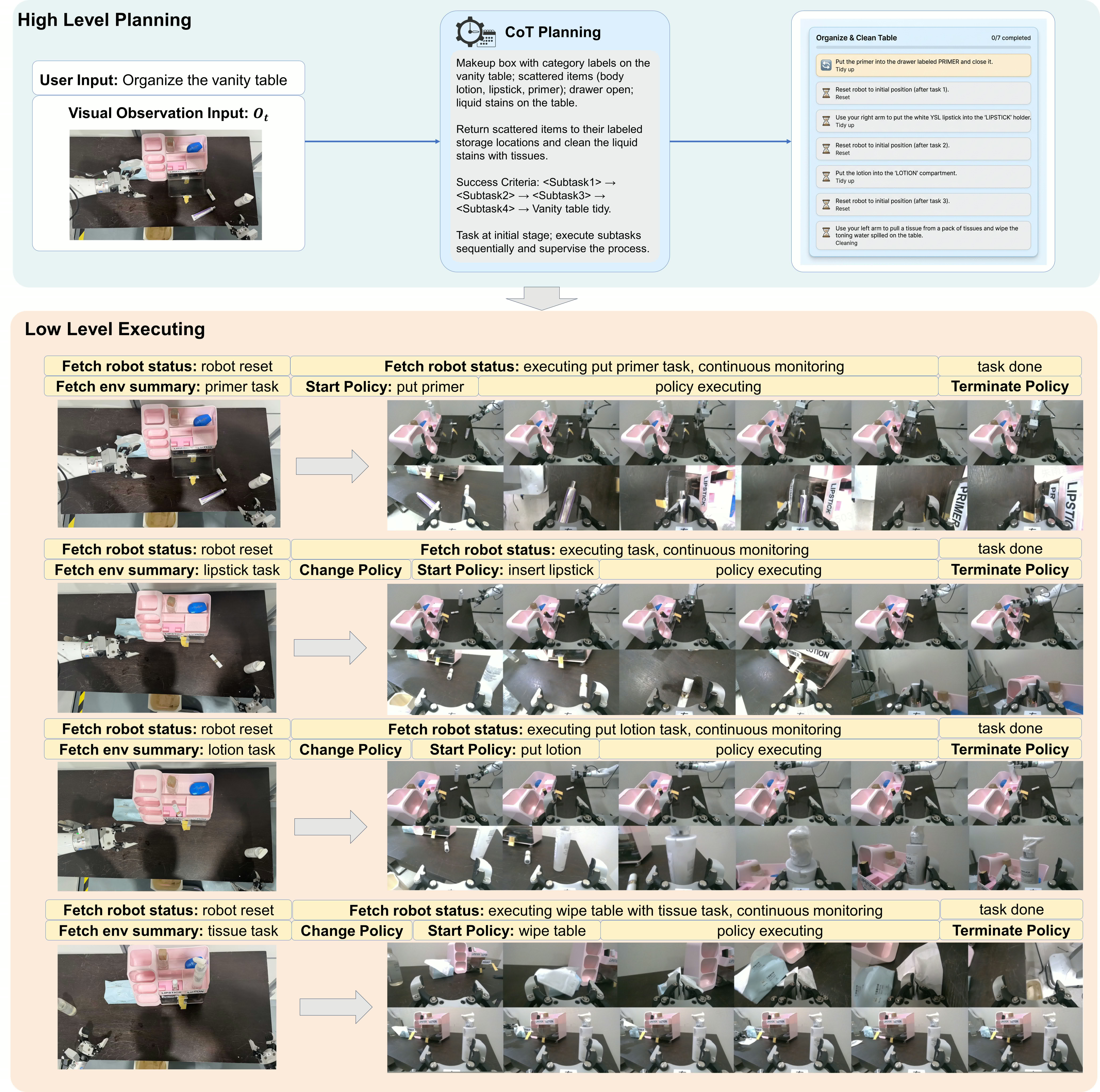}
    \caption{Long-horizon task execution with agent orchestration. The same VLM-based agent plans over the vanity table tidying task and dynamically composes independent forward policy checkpoints (primer placement, lipstick insertion, lotion placement and tissue wipe), invoking re-planning when needed.}
    \label{fig:fig6_long_horizon_demo}
\end{figure}

Finally, we investigate the role of RoboClaw in complex long-horizon tasks. Long-horizon tasks typically consist of multiple sequential steps with dependencies between them, placing higher demands on both planning and execution stability.

To answer Question~3, we evaluate the performance of RoboClaw against two baselines on the vanity table organization task. 
Baseline~1 uses a $\pi_{0.5}$ model trained on the same dataset but without the RoboClaw framework. 
Baseline~2 estimates the expected success rate as the product of the success rates of four subtask policies. 
This comparison allows us to isolate and analyze the contribution of RoboClaw to long-horizon task performance.

The RoboClaw training pipeline incorporates three sources of data: \textbf{a)} human-collected demonstration data, \textbf{b)} autonomously collected RoboClaw trajectories, \textbf{c)} human interventions following failed autonomous rollouts.

The experimental results on the vanity table organization task are shown in Fig.~\ref{fig:human_effort}(c). The results indicate that RoboClaw significantly outperforms both baselines on long-horizon tasks. This improvement comes from RoboClaw’s ability to monitor task progress and automatically invoke recovery policies when necessary.

Figure~\ref{fig:fig6_long_horizon_demo} illustrates a representative execution sequence,
including RoboClaw's planning traces and the sequence of tool invocations
during the vanity table organization task.

\subsection{Can RoboClaw Learn from Failures?}
\label{sec:exp_failure}

During long-horizon execution, RoboClaw summarizes common failure patterns from execution context and interaction history.
Based on these observations, we identify two categories of failures.

\textbf{Non-degrading failures} refer to cases where the environment state remains largely unchanged and the failure can be resolved by retrying the same policy.
For example, during the lotion bottle grasping policy, the gripper may miss the object or close slightly off-target, resulting in an empty grasp.
Since the bottle remains upright and its pose is largely unchanged, the agent can simply retry the same policy without additional recovery actions.

\textbf{Degrading failures} occur when the failure alters the environment state in a way that prevents immediate retry.
For instance, a failed grasp may cause the lotion bottle to tip over or slide away from its initial position, placing it outside the normal precondition region of the grasping policy.
In such cases, additional recovery actions are required to restore a feasible state before the task can continue.

In early rollout stages, such degrading failures often require human intervention to restore the scene.
However, as RoboClaw accumulates execution experience, these recovery behaviors are gradually incorporated into the policy library as dedicated recovery policies.
During later executions, the agent can autonomously invoke these recovery policies to restore the environment and resume the task without human intervention.

This observation suggests that iterative rollout not only improves the robustness of existing policies, but also enables the system to expand its behavioral repertoire by learning recovery strategies.
As the policy library grows to include both nominal policies and recovery behaviors, RoboClaw progressively increases the range of environment states it can reliably handle.

%% file: tables/subtask_reset.tex




\begin{table}[t]
\centering
\caption{Success rates of inverse reset policies across four manipulation tasks.}
\small
\setlength{\tabcolsep}{6pt}
\begin{tabular}{lcccc}
\toprule
\textbf{Task} & \textbf{Body Lotion} & \textbf{Primer} & \textbf{Lipstick} & \textbf{Tissue Wipe} \\
\midrule
Success Rate & 36/50 & 38/50 & 43/50 & 39/50 \\
\bottomrule
\end{tabular}
\label{tab:inverse_skill}
\end{table}

%% file: tables/subtask_effiency.tex
\begin{table}[t]
\centering
\caption{
Success rates of forward manipulation policies across rollout iterations.
}
\small
\setlength{\tabcolsep}{6pt}
\begin{tabular}{ccccc}
\toprule
Iteration & Body Lotion & Primer & Lipstick & Tissue Wipe \\
\midrule
1 & 21/50 & 23/50 & 2/50 & 11/50 \\
2 & 25/50 & 31/50 & 4/50 & 13/50 \\
3 & 32/50 & 31/50 & 11/50 & 14/50 \\
4 & 37/50 & 34/50 & 16/50 & 21/50 \\
5 & 43/50 & 40/50 & 23/50 & 26/50 \\
\bottomrule
\end{tabular}


\label{tab:forward_skill}
\end{table}

%% file: section/conclusion.tex
\section{Conclusion} 
\label{sec:conclusion}

We presented RoboClaw, an agentic robotics framework that unifies data acquisition, policy learning, and long-horizon task execution within a single VLM-driven agent loop. While the framework demonstrates the potential of integrating reasoning, perception, and action within a unified pipeline, it also faces several limitations, including the potential latency introduced by cloud-based large models and the assumption of practical inverse reset behaviors for constructing reusable environment states. Despite these challenges, RoboClaw offers a promising foundation for scalable embodied AI systems. As VLM and VLA models continue to improve, the framework can naturally incorporate stronger models and expand to broader robotic capabilities such as navigation, mobile manipulation, and multimodal interaction, while integrating richer agentic tools for perception, planning, and execution to support more autonomous and adaptable robotic systems.